\documentclass[letterpaper, 10 pt, conference]{ieeeconf}  

\IEEEoverridecommandlockouts                              

\usepackage{amsmath} 
\usepackage{amssymb, gensymb}  
\usepackage{graphicx} 
\usepackage{epsfig} 
\usepackage[capitalise]{cleveref} 
\usepackage{subcaption} 
\newcommand\Ti{\mbox{\textit{Ti}}}  

\title{\LARGE \bf
Improving Swimming Performance in Soft Robotic Fish with Distributed Muscles and Embedded Kinematic Sensing
}

\author{Kevin Soto, Isabel Hess, Brandon Schrader, Shan He and Patrick Musgrave$^{*}$
\thanks{*This material is based upon work that is partially supported by the National Science Foundation under Award No. 2345913}
\thanks{*The authors are with Department of Mechanical and Aerospace Engineering, Herbert Wertheim College of Engineering,
        University of Florida, 1064 Center Dr. Engineering Building Room 181, Gainesville, FL, USA
        {Corresponding Author: \tt\small pmusgrave@ufl.edu}}%
}

\begin{document}

\maketitle
\thispagestyle{empty}
\pagestyle{empty}

\begin{abstract}

Bio-inspired underwater vehicles could yield improved efficiency, maneuverability, and environmental compatibility over conventional propeller-driven underwater vehicles. 
However, to realize underwater vehicles that harness the swimming performance of biology, there is a need for soft robotic swimmers with both distributed muscles and kinematic feedback. 
This study presents the design and swimming performance of a soft robotic fish with independently controllable muscles and embedded kinematic sensing distributed along the body. 
The soft swimming robot consists of an interior flexible spine, three axially distributed sets of HASEL artificial muscles, embedded strain gauges, a streamlined silicone body, and off-board electronics. 
In a fixed configuration, the soft robot generates a maximum thrust of 7.9 mN when excited near its first resonant frequency (2 Hz) with synchronized antagonistic actuation of all muscles. 
When excited near its second resonant frequency (8 Hz), synchronized muscle actuation generates 5.0 mN of thrust. 
By introducing a sequential phase offset into the muscle actuation, the thrust at the second resonant frequency increases to 7.2 mN, a 44\% increase from simple antagonistic activation.
The sequential muscle activation improves the thrust by increasing 1) the tail-beat velocity and 2) traveling wave content in the swimming kinematics by four times.
Further, the second resonant frequency (8 Hz) generates nearly as much thrust as the first resonance (2 Hz) while requiring only $\mathbf{\approx25\%}$ of the tail displacement, indicating that higher resonant frequencies have benefits for swimming in confined environments where a smaller kinematic envelope is necessary. 
These results demonstrate the performance benefits of independently controllable muscles and distributed kinematic sensing, and this type of soft robotic swimmer provides a platform to address the open challenge of sensorimotor control. 

\end{abstract}

\section{INTRODUCTION}

Unmanned underwater vehicles (UUVs) are essential for exploring and monitoring underwater environments that are inaccessible to humans. 
Rigid, propeller-driven UUVs are the standard vehicle for these tasks; however, propeller UUVs are limited due to significant noise generation, limited agility, and the risk of entanglement with vegetation. 
Further, rigid UUVs have poor maneuverability, which presents challenges for operation in dense ecosystems such as the coastal zone. 
These limitations can be addressed by drawing inspiration from biology. 
Fish currently outperform UUVs in terms of agility, indicating that the soft and flexible bodies of fish could yield biologically inspired UUVs with enhanced vehicle performance \cite{fishAdvantagesAquaticAnimals2020}. 
Bio-inspired UUVs would also blend more effectively into the ecosystem to allow for discreet observation of marine life and would minimize environmental damage to yield a more sustainable option for underwater exploration.

Biological swimmers achieve high maneuverability and efficiency through a soft body with distributed muscles and integrated kinematic feedback \cite{smitsUndulatoryOscillatorySwimming2019, jimenezFlexibilityHiddenAxis2023}. 
Fish independently control their axially distributed muscles to yield efficient undulatory swimming \cite{jayneAreMuscleFibers1995} and switch between different swimming modes: actively stiffening their muscles to maintain efficiency across speeds \cite{quinnTunableStiffnessFish2022, jimenezFlexibilityHiddenAxis2023}, swimming with an alternating burst and coast gait to increase efficiency \cite{videlerEnergeticAdvantagesBurstcoast1982, liBurstcoastSwimmersOptimize2021}, or realizing sudden acceleration to maneuver or catch prey \cite{tytellBodyStiffnessDamping2018}. 
Each swimming gait requires unique muscle patterning, and effectively controlling and switching between these gaits requires sensorimotor control that integrates kinematic sensing with distributed muscles. 

To achieve sensorimotor control for adaptation across swimming conditions in manmade systems, robotic swimmers need both controllable muscles groups and distributed kinematic feedback. 
Previous research has successfully demonstrated undulatory soft robotic swimmers \cite{marcheseAutonomousSoftRobotic2014, berlingerModularDielectricElastomer2018, gravertLowvoltageElectrohydraulicActuators2024, hessContinuumSoftRobotic2024}; however, these soft robots typically only have a single muscle group and have no integrated kinematic sensing. 
Swimming using distributed actuators has been demonstrated on eel-like robots with many motor-linkage components \cite{hultmarkFlowfieldMeasurementsWake2007, anastasiadisIdentificationTradeoffSpeed2023}, however, these robots don't take advantage of flexibility and continuum-based bodies which improve swimming performance \cite{quinnTunableStiffnessFish2022}.
More lifelike than motor-driven robots, soft robotic swimmers with independently controllable smart material muscles have been demonstrated \cite{liuMantaRayRobot2022, schwabUndulatorySwimmingPerformance2022, christiansonTranslucentSoftRobots2018}, but kinematic sensing is either absent or limited in spatial fidelity \cite{schwabUndulatorySwimmingPerformance2022}. 
High fidelity kinematic sensing is necessary for robotic swimmers which intend to use sensorimotor control to achieve the diverse swimming modalities seen in biological fish.

This study presents the design and initial testing of a soft robotic swimmer with independently controllable muscles and embedded kinematic sensing distributed along the body. 
The presented soft robot is a slender-bodied undulatory swimmer with three axially distributed groups of hydraulically amplified, self-healing electrostatic (HASEL) artificial muscles, and strain gauges embedded along the body. 
This paper presents the swimming performance for the soft robotic swimmer in quiescent water, in a mounted configuration and with offboard electronics. 
We quantify the thrust and kinematic performance of the soft robotic swimmer for different muscle excitation frequencies and phases. 
This study demonstrates that independent control of the distributed muscle phasing can yield improved thrust and undulatory kinematics. 
This soft robotic swimmer with distributed muscles and sensing will serve as a platform to address the open challenges of sensorimotor control and will enable promising machine learning frameworks that harness the embodied intelligence of the swimmer \cite{hePhysicalReservoirComputing2025}.

\section{DESIGN AND FABRICATION OF ROBOTIC SWIMMER} \label{sec: methods}

\begin{figure}
    \centering
    \begin{subfigure}{\linewidth}
        \centering
        \includegraphics[width=\linewidth]{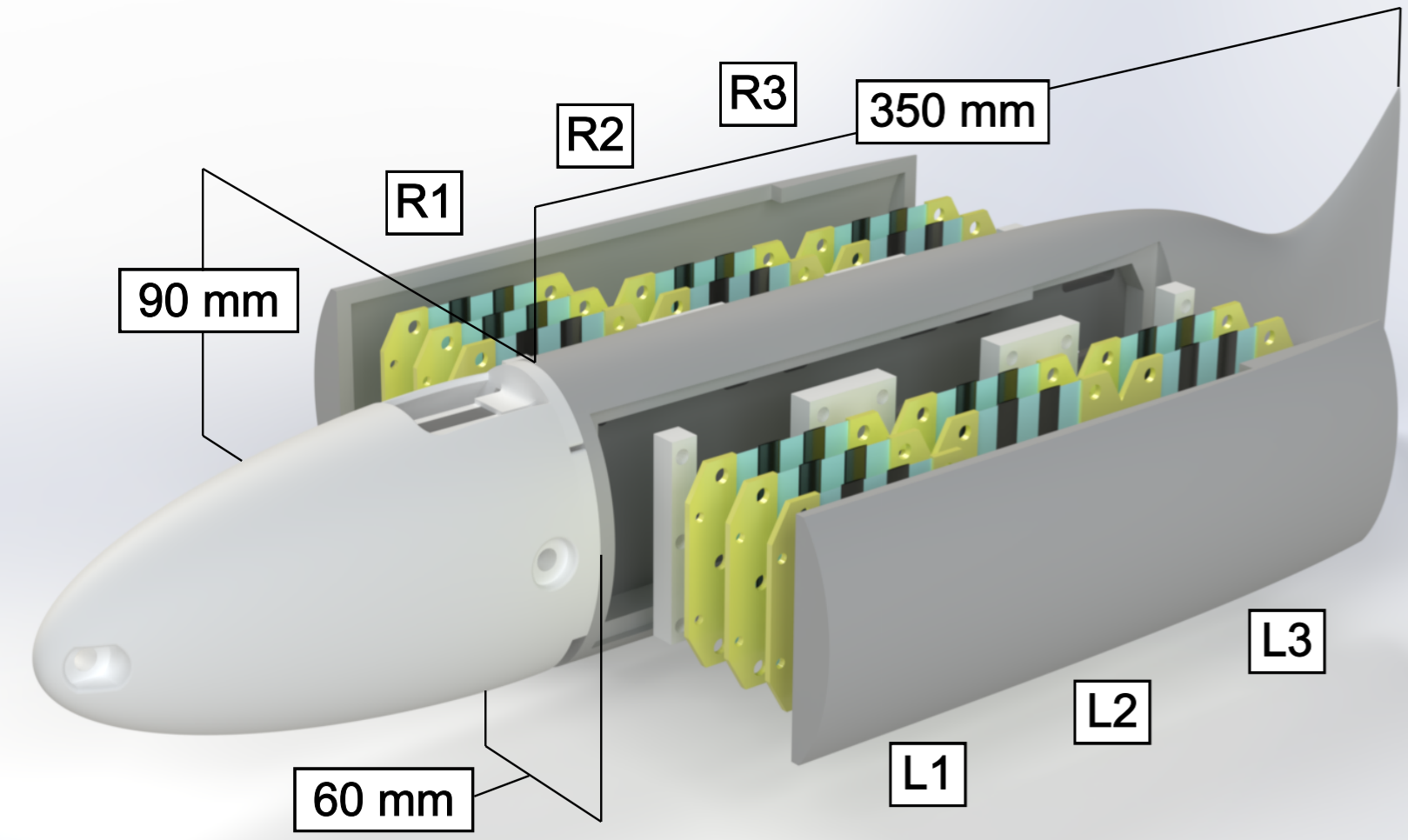}
        \caption{CAD rendering.}
        \label{fig: cad rendering}
    \end{subfigure}
    \hfill
    \begin{subfigure}{\linewidth}
        \centering
        \includegraphics[width=\linewidth]{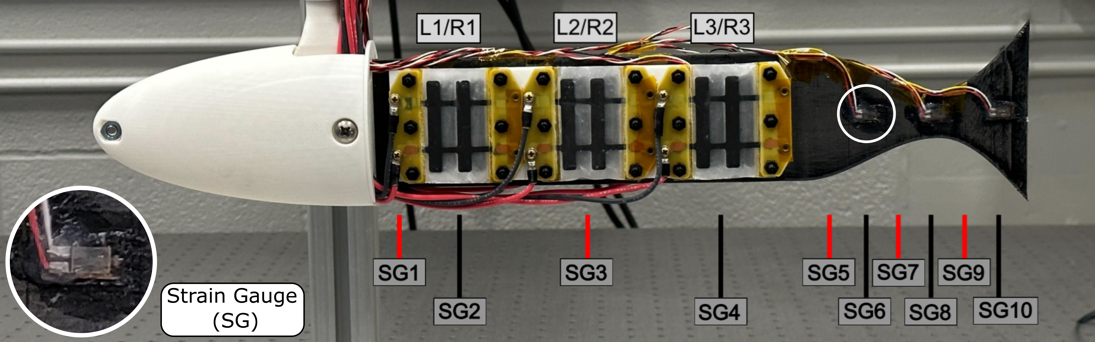}
        \caption{Before silicone molding.}
        \label{fig: fabrication before silicone}
    \end{subfigure}
    \hfill
    \begin{subfigure}{\linewidth}
        \centering
        \includegraphics[width=\linewidth]{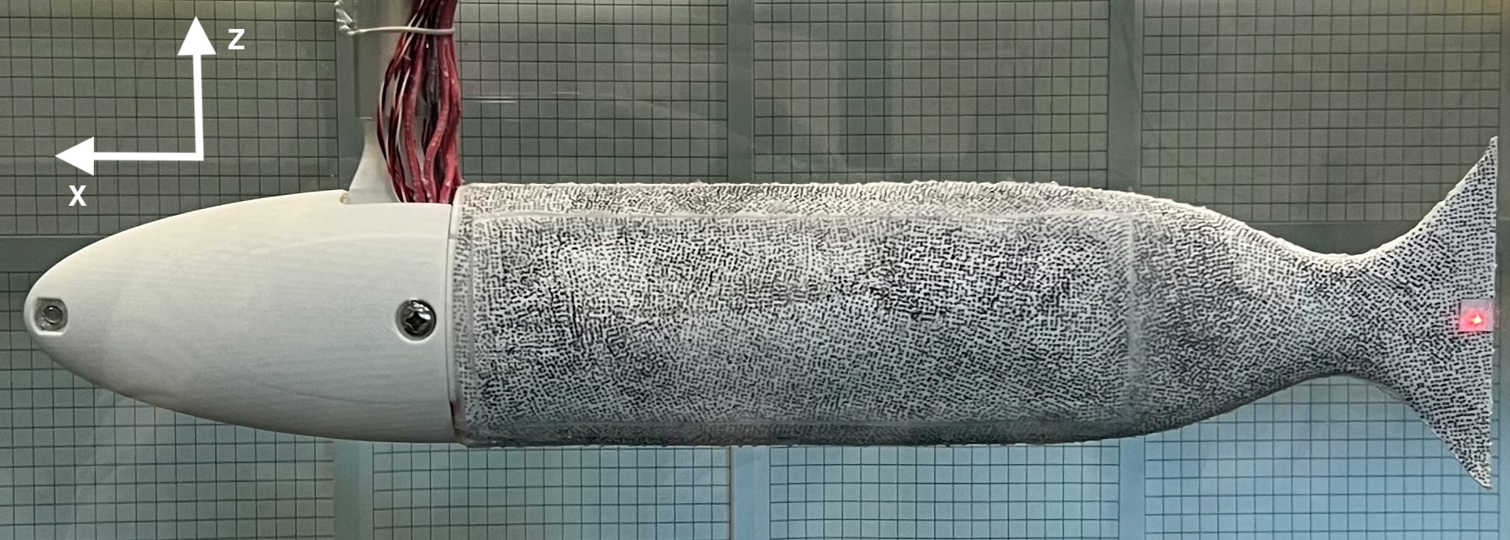}
        \caption{Completed robot with silicone cover attached.}
        \label{fig: fabrication complete}
    \end{subfigure}
    \hfill
    \caption{Fabrication process. (a) Exploded CAD rendering displaying internal structure and major dimensions, omitting wiring and mounting hardware. (b) Before silicone molding showing internal skeleton and HASEL locations.  Left-side mounted strain gauges are even numbered, right-side are odd numbered. All are attached along the robot's midline. (c) Complete robot submerged underwater.}
    \label{fig: fabrication}
    \vspace{-1.5em}
\end{figure}

The soft robotic fish was designed with six distributed muscle groups and integrated strain sensing along the length.
The robot's geometry was optimized through an automated parameter study using a previously validated kinematic model of a soft robotic swimmer \cite{hessContinuumSoftRobotic2024}. 
The objective was to maximize the tail-beat velocity, which would subsequently maximize thrust according to the large-amplitude elongated body theory of undulatory propulsion \cite{lighthillLargeAmplitudeElongatedBodyTheory1971}. 
The parameter study varied the caudal fin span and chord, the location and width of the peduncle joint, the body length, width profile, and thickness profile along the length.
The optimized soft robotic swimmer is shown in \cref{fig: fabrication} and consists of three primary components: a flexible spine assembly with the actuators and strain gauges, a molded silicone body, and a rigid head for mounting. 
This section presents the fabrication of the optimized design.

The spine assembly consists of a flexible TPU spine, 14 HASEL actuators, and ten strain gauges.
The robot's skeleton has a body length (bL) of 350 mm (X direction) and was constructed with 95A durometer thermoplastic polyurethane (TPU) by fused deposition modeling. 
The spine is 7 mm thick (Y direction) from the head to the the caudal fin ($x=0.64$ bL), after which it tapers linearly to 1 mm at the tail tip. 
The width (Z direction) profile consists of a 70 mm wide rectangle through the actuated region, narrows to 26 mm at the peduncle joint ($x=0.85$ bL), and grows to 100 mm at the tail tip. 

HASEL actuators are soft artificial muscles which contract with high voltage \cite{acomeHydraulicallyAmplifiedSelfhealing2018}.
HASEL actuators were chosen as the artificial muscles for this platform because they have force output, stroke length, and frequency bandwidth comparable to biological muscles \cite{rothemundHASELArtificialMuscles2021}.
Fourteen actuators are distributed along the robot's body in an antagonistic configuration, with seven per side, across three axial locations shown in \cref{fig: cad rendering}.
The anterior positions nearest to the head (L1/R1) have three HASELs each on the left and right side of the spine. The posterior two positions (L2/R2 and L3/R3) were each fitted with two HASELs per side. 
More HASELs were employed toward the head because the head creates a clamped condition on the body, and Euler-Bernoulli beam theory states that greater moments near the head yields greater caudal fin deformation.
The actuators are secured to the skeleton through slots in the TPU spine.
Plastic standoffs distance the HASELs 8 mm from the spine, both to provide clearance for the HASELs to expand and to magnify the bending moment generated on the spine by their contraction. 

Strain gauges were embedded in the robot to provide distributed kinematic sensing while in motion.
Ten $0.25$in long, metal foil strain gauges (Micro Measurements MMF002564) are glued to the TPU spine along the body length as shown in \cref{fig: fabrication before silicone}.
They are split evenly between the left and right sides, five per side, alternating along the length.
Only the anterior six of the ten strain gauges are used for kinematic sensing in this study; the four gauges nearest to the tail (SG 7-10) failed during testing due to insufficient strain relief.  

The spine assembly was cast in a silicone body to provide a streamlined fish shape. The silicone body was designed with an internal cavity (Figure \ref{fig: cad rendering}) to prevent the silicone from restricting the HASEL's contraction, increasing their effectiveness from previous robots \cite{hessContinuumSoftRobotic2024}.
After the strain gauges were attached, the TPU skeleton was cast in silicone leaving a cavity for the HASELs to be reattached. 
The silicone is a 23:1 mixture of Ecoflex 00-10 (Smooth On, Inc.) silicone to 3M glass bubbles S22 by mass, achieving a density of 0.91 g/cm$^3$. The glass bubbles were added to improve the robot's buoyancy.
Once the silicone was cured, the HASELs were secured to their designated locations and tensioned.

Silicone covers were secured over the rectangular cavity to seal the HASELs into the robot. The mixture ratio of silicone to glass bubbles for the covers was 10:1 by mass, achieving a density of 0.76 g/cm$^3$. 
These covers were then bonded to the body using Sil-Poxy (Smooth On, Inc.). 
FR3 dielectric oil was injected into the body cavity to provide an insulating substrate for the HASELs and maintain the body's streamlined shape.
The complete robot is shown in \cref{fig: fabrication complete}. The exterior is painted with a speckle pattern for follow-on digital image correlation experiments. The complete body has a varying elliptical cross-section designed from the parameter study. The width (Z direction, major axis) of the silicone body tapers from 90 mm to 40 mm at the peduncle joint, and grows to 120 mm at the caudal fin trailing edge. The thickness (Y direction, minor axis) of the silicone body tapers from 60 mm to 5 mm at the trailing edge. 

The spine and silicone body are attached to a rigid internal mounted component which interfaces the rear elastic portion of the body with the streamlined rigid head. The external mounting point holds the fish fixed during testing and provides a clamped condition for the elastic body. 
The robot's rigid head has a modular design so that accessories can be mounted after the body has been manufactured. 
Future iterations of the head will be designed for free swimming capability.

\section{EXPERIMENTAL METHODS}
\label{sec: exp setup}
\begin{figure}
    \centering
    \includegraphics[width=\linewidth]{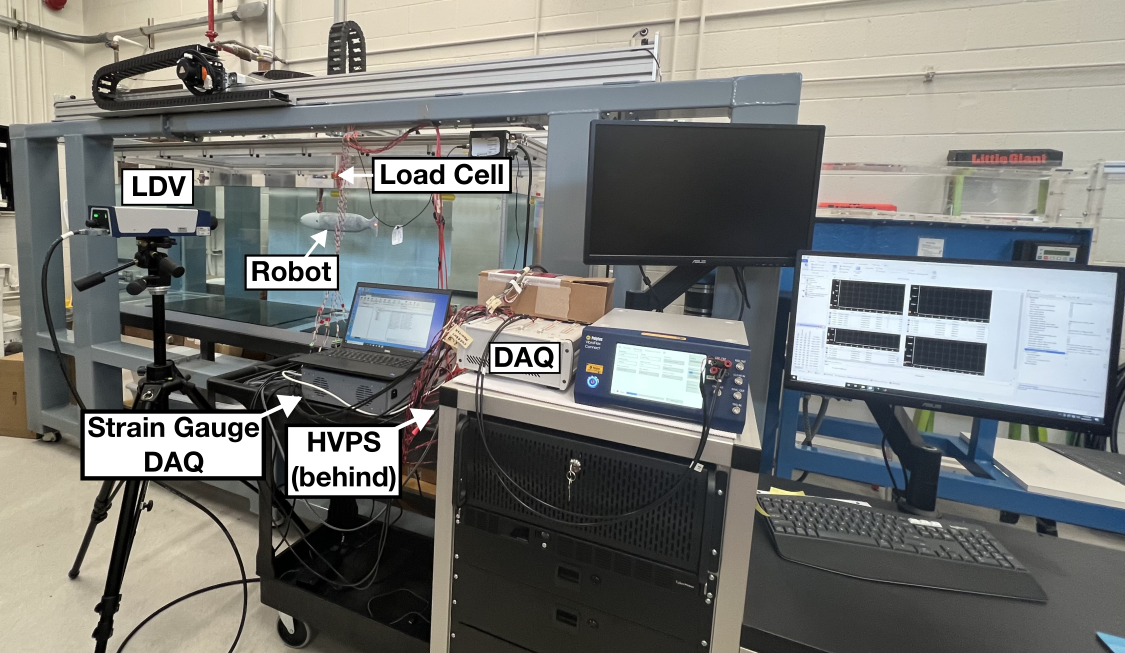}
    \caption{Experimental setup with labeled equipment.}
    \label{fig: setup}
    \vspace{-1.5em}
\end{figure}

The kinematic and thrust performance of the soft robotic swimmer were experimentally quantified in quiescent water using the experimental setup shown in \cref{fig: setup}. 
The experimental testing has two aims: 
1) determine the submerged swimmer's frequency-dependent kinematics and natural frequencies by experimental modal analysis and
2) quantify the robot's thrust and continuous-body swimming kinematics under harmonic excitation of the HASEL actuators.

All tests were conducted in an 8 ft $\times$ 3 ft $\times$ 3 ft tank of quiescent tap water. A six-axis load cell (ATI Industrial Automation Mini40) was installed in line with the rigid mount at the head to measure the reaction forces and moments generated by the robotic fish.
The strain gauges embedded in the robot were used to recreate the midline kinematics of the soft body's swimming motion. A Micro Measurements 9000-16-SM-AO StrainSmart DAQ was used in quarter bridge configuration with 8 V excitation and the strain signals were filtered with a 125 Hz low-pass filter. In complement, a single-point Polytec VibroFlex Neo VFX-I-110 laser Doppler vibrometer (LDV) gathered high fidelity velocity measurements at the caudal fin trailing edge. The tail-tip was chosen as its motion plays the largest role in the fish's swimming performance \cite{lighthillLargeAmplitudeElongatedBodyTheory1971}. 
All signals were recorded with a dSpace MicroLabBox at 1 kHz sampling rate.

Each of the six sets of muscles along the fish's length were independently controlled to study varying the phase offsets between muscles along the length.
Unipolar, 6 kV sinusoidal muscle activation signals up to 9 Hz are investigated in this study. 
The high voltage power signals were supplied to the HASEL groups using an Artimus Robotics PS2-08-A-03 high voltage power supply.

The thrust was calculated from the reaction forces and moments measured by a six-axis load cell on the rigid mounting post. 
The force in the axial direction, $F_X$, oscillates with the caudal fin's motion during steady state swimming and the net thrust is the moving mean of the force. Since the force is periodic, taking the moving mean over an integer number of periods reveals the thrust, but that is sensitive to precisely knowing the oscillation period and the sampling rate aligning with the oscillation frequency. Instead, we calculated the time-varying thrust through convolution of $F_X$ with a four-period-wide Blackman-Harris window, normalized by the width of the window, to reduce the effects of the edge of the averaging window.

Finally, the swimming waveform during steady state oscillation was reconstructed from the embedded strain gauge data and the caudal fin displacement measured with the LDV. 
The displacement was calculated using the relationship between strain and deflection in an Euler-Bernoulli beam, \eqref{eq: deflection} \cite{inmanEngineeringVibration2008}.
The robot's strain profile, $\varepsilon(x,t)$, was represented by spline interpolation of the strain gauge data. The curvature was calculated by dividing the strain profile by the strain gauge's distance from the neutral axis, $z=t/2=3.5\ \textrm{mm}$. Then, the body deflection, $w(x,t)$, was calculated by integrating the curvature twice over the length, where $\chi_1$ and $\chi_2$ are dummy variables of integration. The measured caudal fin displacement from the LDV provided the deformation at the trailing edge.

\begin{equation}
    w(x,t) = \int_{\chi_1=0}^{\chi_1=x} \int_{\chi_2=0}^{\chi_2=\chi_1} \frac{\varepsilon(\chi_2,t)}{z}\,\textrm{d}\chi_2 \,\textrm{d}\chi_1
    \label{eq: deflection}
\end{equation}

The body deflection over time, $w(x,t)$, is the robot's swimming kinematics. The oscillation can be a standing wave, a traveling wave, or some complex wave with both standing and traveling components. The proportion of traveling waves in the body waveform over time was quantified using the traveling index \cite{feenyComplexOrthogonalDecomposition2008, musgraveElectrohydroelasticModelingStructureBorne2021}. 
The traveling index, $\Ti$, is a complex orthogonal decomposition method which reports a scalar value corresponding to the proportion of traveling to standing waves: $\Ti = 1$ is a pure traveling wave, $\Ti = 0$ is a pure standing wave, and $0 < \Ti < 1$ is a complex wave.
The traveling index was computed for the robot's deflection measured by the strain gauges over one period of steady state oscillation.
The envelope of the swimming mode is defined as the maximum and minimum of the deflection at each point along the length over time.

\section{RESULTS}
\label{sec: results}
The swimming performance of the robotic fish measured using the experimental setup described in \cref{sec: exp setup} is presented in this section. 
First, the damped natural frequencies of the soft robot are presented to identify operating frequencies.
Then, the thrust and swimming kinematics are presented under harmonic excitation of the rigidly tethered swimmer at the operating frequencies with varied muscle phasing. 

\subsection{Natural Frequencies}
\begin{figure}
    \centering
    \includegraphics[width=1\linewidth]{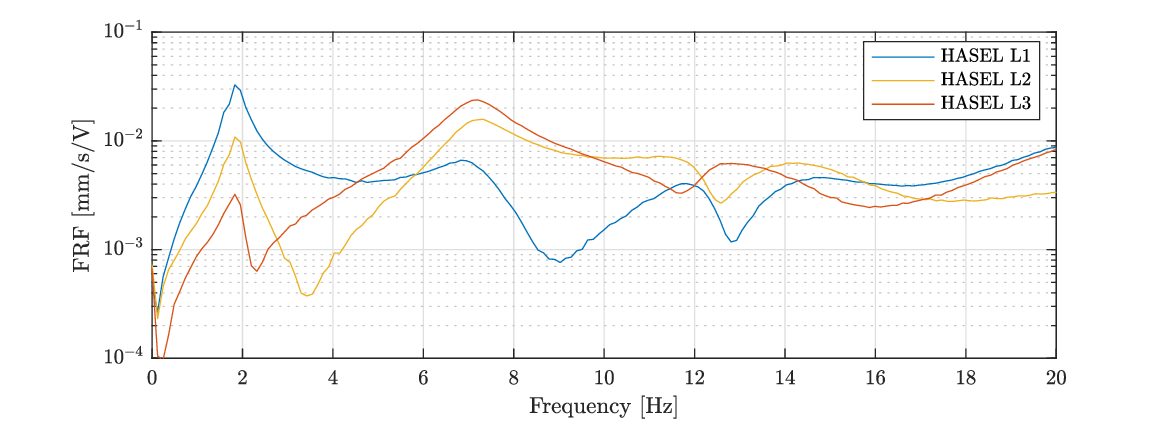}
    \caption{Frequency response of the tail-tip velocity (normalized by input voltage) for each HASEL group.}
    \label{fig: impulse H2}
        \vspace{-1.2em}
\end{figure}

The damped natural frequencies of the submerged swimmer were found by an impulse response test.
An electrical impulse was sent to each HASEL group shown in \cref{fig: fabrication before silicone} to determine the submerged robot's velocity frequency response function (FRF). The input-normalized FRFs were averaged over four repetitions for each actuator group. The response for HASEL groups L1, L2, and L3 are shown in \cref{fig: impulse H2}; where excitation by groups R1, R2, and R3 would yield similar responses due to symmetry. 
The swimmer's lowest two damped natural frequencies are 1.9 Hz and 7.0 Hz and correspond  with the peaks in the FRF. 
Due to the differing axial locations, each muscle group yields a different frequency-dependent tail response; however, the natural frequencies are a global property of the swimmer and are the same for all muscle groups. 

\subsection{Swimming Performance}
The robot's swimming performance was quantified by measuring the thrust and kinematics under harmonic excitation of the distributed muscles while varying input frequency and muscle phasing. 
The thrust and operational deflection shape (swimming kinematics) were characterized around the damped natural frequencies. The cases which produced the largest thrust are discussed in this section. The results are summarized in \cref{tab: results}.

\subsubsection{In-Phase Actuation}
\begin{figure}
    \centering
    \begin{subfigure}{\linewidth}
        \centering
        \includegraphics[width=1\linewidth]{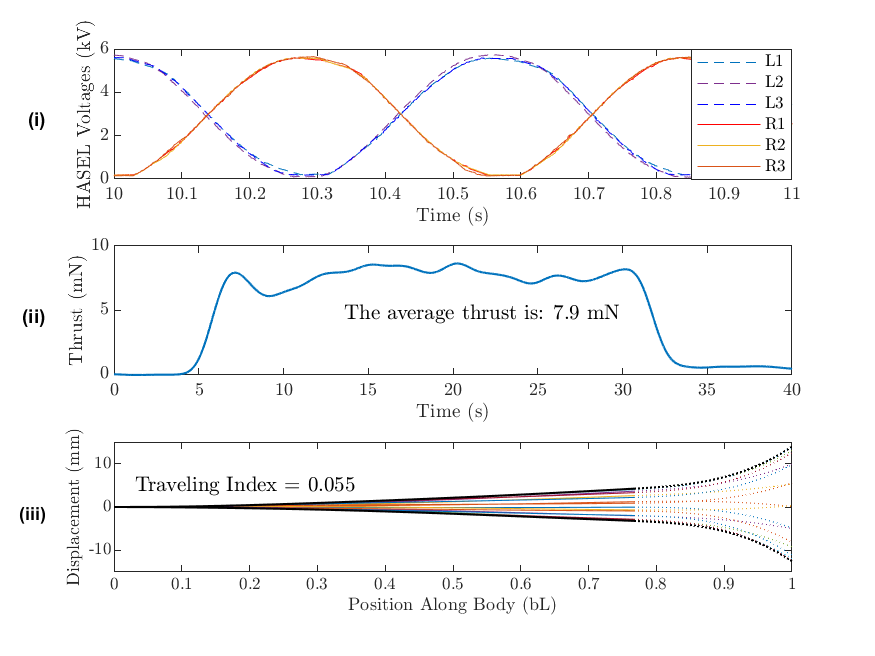}
        \caption{2.05 Hz. The deflection strobe lines are 36 ms apart.}
        \label{fig: 2.05 result}
    \end{subfigure}
    \hfill
    \begin{subfigure}{\linewidth}
        \centering
        \includegraphics[width=1\linewidth]{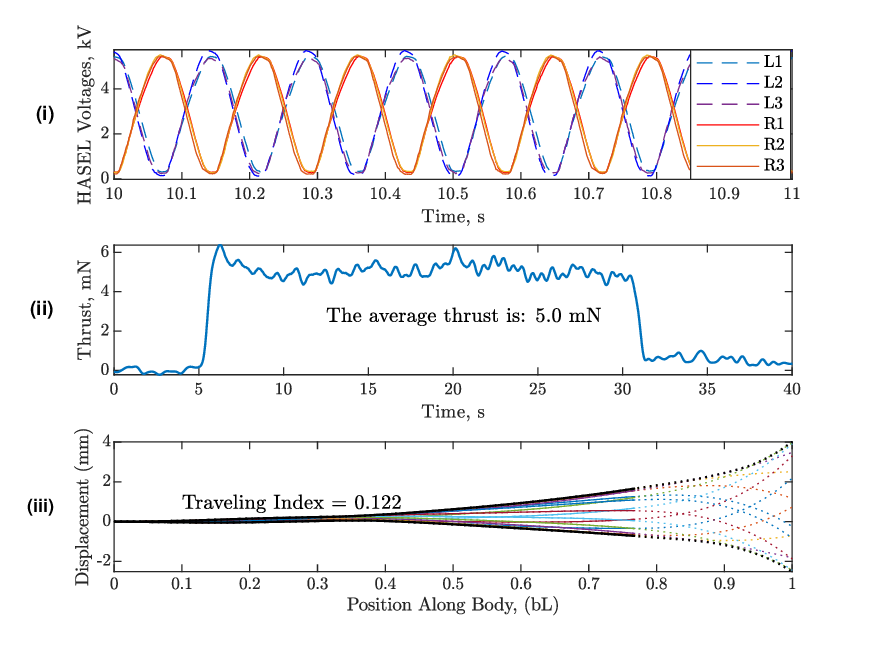}
        \caption{8.05 Hz. The deflection strobe lines are 9.1 ms apart.}
       \label{fig: 8.05 result}
    \end{subfigure}
    \hfill
    \caption{Swimming performance for in-phase antagonistic actuation. Each response (a) 2.05 Hz and (b) 8.05 Hz are broken into three parts: Muscle excitation waveform (i), time varying thrust (ii), and spatially varying swept envelope (iii) for in-phase cases. The window for the muscle excitation waveform is trimmed to show one second of oscillation. The time varying thrust plot's window is over the length of an entire trial. The swept envelope is shown over one period of oscillation with 16 linearly spaced strobe lines. }
    \label{fig: no offset result}
    \vspace{-1em}
\end{figure}

The robot's swimming performance was first characterized with simple antagonistic actuation. 
\cref{fig: no offset result} displays the swimming performance of the robot at its greatest thrust producing frequencies, $f_1=2.05$ Hz and $f_2=8.05$ Hz. The operational frequencies with the largest thrust, $f_1$ and $f_2$, are slightly higher than the damped natural frequencies which have the greatest tail-beat velocity. This difference between the resonant frequency and the frequency with the greatest thrust is likely due to nonlinear fluid loading effects. 

We define simple antagonistic actuation as a pattern where all the HASELs on the robot's left side (L1-L3) are contracted together and alternate with the HASELs on the right side (R1-R3) with $180\degree$ phase offset. 
The antagonistic voltage excitation to the six HASEL muscles is shown in \cref{fig: 2.05 result} (i) and \cref{fig: 8.05 result} (i).

The thrust over the entire test period, shown in \cref{fig: 2.05 result} (ii) and \cref{fig: 8.05 result} (ii), is the average once the swimmer reaches steady state. The steady state thrust was 7.9 mN at $f_1=2.05$ Hz and 5.0 mN at $f_2=8.05$ Hz with simple antagonistic actuation.

The kinematic response is shown in \cref{fig: 2.05 result} (iii) and \cref{fig: 8.05 result} (iii), where the black lines are the displacement envelopes and the colored lines are a strobe of the displacement at 16 instances over a single period. 
The displacement in $x/L \leq 0.75\ \textrm{bL}$ (solid lines) were generated using the strain gauges, the displacement at the tail tip is the LDV measurement, and the displacement in $x/L > 0.75\ \textrm{bL}$ (dashed lines) were generated by a spline interpolation between the calculated displacement and the LDV data. 
Simple antagonistic actuation produced a kinematic response with low traveling index, $\Ti=0.055$ at $f_1$ and $\Ti=0.122$ at $f_2$, indicating the swimming mode was standing wave dominant. 

t

The caudal fin velocities under simple antagonistic actuation at $f_1$ and $f_2$ are very similar, 150 mm/s and 148 mm/s respectively, but the two frequencies produce different amounts of thrust. 
This is likely because they have different absolute deflection magnitudes. The caudal fin deflection at $f_1$ was $|w_{f_2}(L,t)|=\pm 13.2$ mm whereas at $f_2$ the displacement is four times smaller at $|w_{f_2}(L,t)|=\pm 3.25$ mm. 
This difference changes the how the fluid interacts with the robot, the large deformation at $f_1$ moves more fluid mass per stroke than smaller tail-beats, and thus increases the thrust produced.

\subsubsection{Sequential Actuation}
\begin{figure}[t]
    \centering
    \includegraphics[width=1\linewidth]{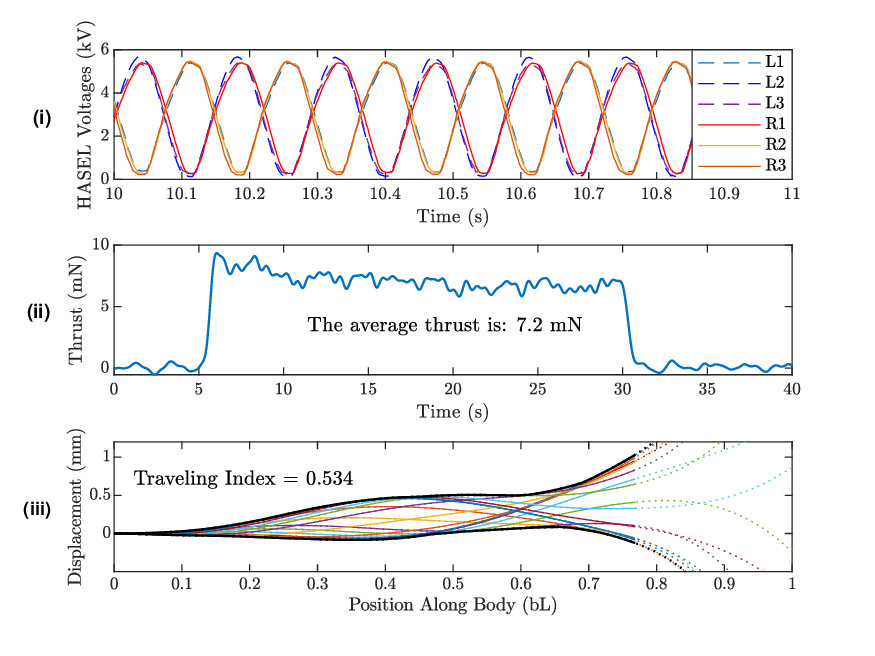}
    \vspace{-1em}
    \caption{Swimming performance for out-of-phase case: 8.05 Hz, $180\degree$ paired offset. Muscle excitation waveform (i), time varying thrust (ii), spatially varying swept envelope (iii). The deflection strobe lines are 9.1 ms apart. }
    \label{fig: 8.05 result offset}
    \vspace{-1em}
\end{figure}

After analyzing simple antagonistic actuation, we examined the role of phase offsets between the muscle groups. 
The configuration with the best performance excited muscle groups 2 and 3 $180\degree$ out of phase from group 1, and maintained antagonistic actuation between the left and right sides.
The swimming performance with this pattern actuated at $f_2=8.05$ Hz is shown in \cref{fig: 8.05 result offset}.  
The soft robot produced 7.2 mN of thrust with a moderate amount of traveling wave content: $\Ti=0.53$. 
This level of traveling wave content is within the range used by biological trout (0.52-0.78) \cite{cuiComplexModalAnalysis2018} and is an improvement over previous continuum soft robotic swimmers \cite{hessContinuumSoftRobotic2024}.
Introducing this axial phase offset significantly improved swimming performance across all metrics: increasing traveling wave content by 430\%, thrust by 44\%, and tail-tip velocity by 14\%, compared to its in-phase counterpart.

With 7.2 mN of thrust, the robot could achieve an estimated free swimming velocity of 0.39 bL/s assuming turbulent flow conditions.
This free swimming velocity is approximated by balancing the thrust with estimates of the skin friction drag and profile drag: $T=\frac{1}{2}{\rho}{v}^2(C_{f}A_{wetted}+C_{d}A_{cross})$, where $C_f$ comes from Prandtl's one-seventh-power law, and $C_d=0.017$ equals that of a rainbow trout which has a similar hydrodynamic profile to the robot \cite{sagongHydrodynamicCharacteristicsSailfish2013}.

\cref{fig: 8.05 result offset} shows that the kinematics have a slight asymmetry that yields a positive static displacement. This asymmetry corresponds with a $0.1^\circ$ deflection of the neutral axis versus the body length; the scaling of the figure makes the asymmetry appear larger. 
This minor asymmetry likely stems from slight differences in the stiffness and HASEL strength on either side of the robot's mid-plane. 
During free-swimming, this minor asymmetry could be corrected by the swimmer's fins which is standard procedure for all biological fish and underwater vehicles.

\begin{table*}
\caption{Swimming Performance Results by Actuation Mode} \label{tab: results}
\centering
\begin{tabular}{|c|c|c|c|c|c|c|}
\hline
Actuation Mode & Frequency, Hz & Muscle Phasing & Thrust & Caudal Deflection & Caudal Velocity & \Ti \\
& (Hz) & (deg) & (mN) & (mm) & (mm/s) & \\
\hline
In-phase $f_1$ &2.05 & $\begin{matrix} L1=0 & L2=0 & L3=0\\ R1=180 & R2=180 & R3=180 \end{matrix}$ & 7.9 & 13.2 & 150 & 0.055\\
\hline
In-phase $f_2$&8.05 & $\begin{matrix} L1=0 & L2=0 & L3=0\\ R1=180 & R2=180 & R3=180 \end{matrix}$ &5.0 & 3.25 & 148 &0.122\\
\hline
Sequential $f_2$&8.05 & $\begin{matrix} L1=180 & L2=0 & L3=0\\ R1=0 & R2=180 & R3=180 \end{matrix}$ &7.2 & 4.05 & 169 &0.534\\
\hline
\end{tabular}
\vspace{-.5em}
\end{table*}

The phase offset improves the thrust by increasing both 1) the tail-tip velocity and 2) the traveling wave content. 
Per the large-amplitude elongated body theory (LAEBT), the thrust generated by a slender undulatory swimmer is proportional to the square of the caudal fin's transverse velocity $T\propto v^2|_{x=L}$ \cite{lighthillLargeAmplitudeElongatedBodyTheory1971}.
By introducing an axial phase offset for excitation at $f_2=8.05Hz$, we increased the caudal fin velocity from 148 mm/s to 169 mm/s, which is a 14\% increase. 
Per the LAEBT, this 14\% increase in caudal fin velocity should yield a 30\% increase in thrust. 
However, we experimentally measured a 44\% increase in thrust. Thus, the improvement in thrust is not only due to the increase caudal fin velocity, which indicates that the increase in traveling wave content also contributes to the improved performance (Table \ref{tab: results}).
This result agrees with previous studies, which have shown that traveling waves can increase the swimming performance due to the swimmers ability to better engage the fluid \cite{ramananarivoPassiveElasticMechanism2013}.

\section{CONCLUSIONS}

This study presents the swimming performance of a soft swimming robot with independently controllable muscles and embedded kinematic sensing distributed along the length. 
The swimming robot consists of an elastic body with 14 HASEL artificial muscles distributed into six independently controllable muscle groups along three axial positions.
Kinematic sensing was achieved with metal foil strain gauges embedded along the length which were used to reproduce the robot's elastic deformation during swimming motion.
The robot's swimming kinematics and thrust were captured in quiescent water for harmonic muscle excitation around the first and second damped natural frequencies. The effect of introducing a phase offset in muscle activation along the length was investigated. 

Independent, sequential control of the muscle groups improved thrust output by 44\% when actuating near the second damped natural frequency.
Phased muscle activation also improved the biomimicry of the swimming waveform by inducing more traveling waves. The traveling index increased from 0.12 to 0.53 with the addition of the phase offset, which puts the motion in range of biological swimmers.
This thrust increase is due to a combination of increased tail tip velocity and greater traveling wave content. 

These results demonstrate the importance of 1) independently controllable muscles in the swimming performance and 2) the role of kinematic sensing in quantifying the swimming performance. 
Future work will identify the mechanistic relationships governing phased muscle actuation, determine the performance under free swimming, and integrate the kinematic sensing for feedback control.
This results and the developed platform will advance the design of bio-inspired unmanned underwater vehicles that can harness the benefits of biological swimmers to improve the swimming performance compared to conventional propeller-driven vehicles.

\addtolength{\textheight}{-12cm}  







\bibliographystyle{ieeetr}
\bibliography{Zotero-Library-bibtex}

\end{document}